\pdfoutput=1

\documentclass[11pt]{article}

\usepackage[final]{acl}

\usepackage{times}
\usepackage{latexsym}
\usepackage{adjustbox}
\usepackage[T1]{fontenc}
\usepackage{listings}

\usepackage[utf8]{inputenc}
\usepackage{booktabs}
\usepackage{microtype}
\usepackage[utf8]{inputenc}

\usepackage{inconsolata}
\usepackage{tabularray}
\usepackage{CJKutf8}

\usepackage{graphicx}
\usepackage{amsmath}
\usepackage{array}

\title{Re:Member: Emotional Question Generation from Personal Memories}

\author{
 \textbf{Zackary Rackauckas\textsuperscript{1}},
 \textbf{Nobuaki Minematsu\textsuperscript{2}}
 \textbf{Julia Hirschberg\textsuperscript{1}},
\\
 \textsuperscript{1}Columbia University,
 \textsuperscript{2}The University of Tokyo,
\\
 \small{
   \textbf{zcr2105@columbia.edu, mine@gavo.t.u-tokyo.ac.jp, julia@cs.columbia.edu}
 }
}

\begin{document}
\maketitle
\begin{abstract}
    We present Re:Member, a system that explores how emotionally expressive, memory-grounded interaction can support more engaging second language (L2) learning. By drawing on users’ personal videos and generating stylized spoken questions in the target language, Re:Member is designed to encourage affective recall and conversational engagement. The system aligns emotional tone with visual context, using expressive speech styles such as whispers or late-night tones to evoke specific moods. It combines WhisperX-based transcript alignment, 3-frame visual sampling, and Style-BERT-VITS2 for emotional synthesis within a modular generation pipeline. Designed as a stylized interaction probe, Re:Member highlights the role of affect and personal media in learner-centered educational technologies.

\end{abstract}

\begin{figure*}[h]
    \centering
    \includegraphics[width=\linewidth]{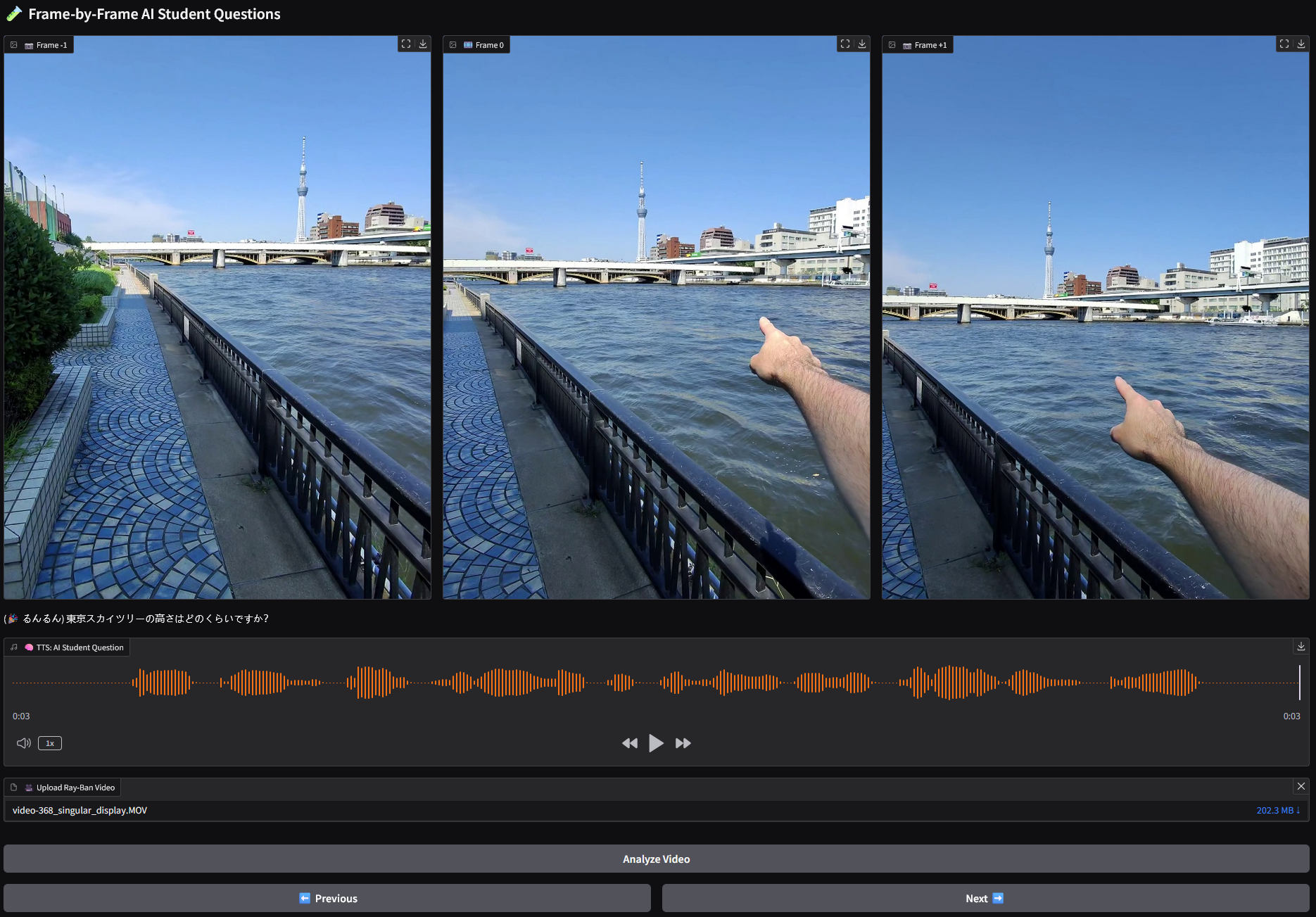}
    \caption{Example interface frame from video (1), showing (from top to bottom) three frames of sequential visual context, the generated emotion, the generated system question, a playable text-to-speech box, the name of the video file, the "Analyze Video" button, and "Previous" and "Next" buttons to navigate between sequential moments.}
    \label{fig:video1}
\end{figure*}

\section{Introduction}
As language learning technologies evolve, there is growing interest in systems that go beyond rote vocabulary drills or disembodied text. Research in Human–Computer Interaction (HCI) and Natural Language Processing (NLP) has shown that social presence, emotional involvement, and personal relevance significantly improve learning outcomes, especially for the acquisition of second languages (L2). However, most existing tools are based on generic or de-contextualized content, limiting their potential to tap into the emotional and mnemonic power of a learner’s lived experiences.

A growing body of HCI research has explored how large language models (LLMs) can support learners and designers through agent-assisted creativity. Systems such as IdeationWeb \cite{10.1145/3706598.3713375} and Promptify \cite{brade2023promptifytexttoimagegenerationinteractive} scaffold user interaction with generative models, enabling iterative refinement and analogical exploration. In language learning, voiced chatbot interfaces, such as conversational characters and stress-free conversational partners \cite{rackauckas2025learningjapanesejouzuinteraction, aiba24_interspeech}, have shown how conversational systems can support learners by tailoring responses to their needs. Related work in agent-assisted creativity and co-design highlights the importance of aligning model outputs with user intent and emotional framework \cite{shaer2024groupaibrainwriting, sun2025integratingemotionalintelligencememory}.

From an NLP perspective, recent work on question generation has moved toward more context-sensitive and user-aligned output. Newer methods leverage LLMs for conversational foresight \cite{guo-etal-2024-pcqpr} and empathetic dialogue \cite{li-etal-2024-eden} where the user's inferred state shapes responses for empathy and engagement \cite{rashkin-etal-2019-towards}. Our work contributes to this work by combining environmental-aware inference from sequential visual frames with LLM-based question generation in a real-time learner interface. The system supports reflective learning by surfacing system-generated, context-sensitive questions that adapt to the learner's evolving affective and attentional state, a goal aligned with broader calls for emotionally intelligent educational technologies \cite{darling-hammond2017effective}. This bridges recent work in HCI and NLP on responsive, learner-aware systems for mixed-initiative interaction.

Our system builds on previous work by grounding LLM-generated questions in video-based emotion cues, enabling emotionally responsive interactions that match the learner’s current context. Specifically, we present Re:Member\footnote{https://github.com/zackrack/Re-Member}, an open-source system that turns videos of personal memories, also known as episodes, such as casual recordings of travel, family, or everyday life, into emotionally voiced, interactive prompts for language learning. By combining recent advances in large language models (LLMs), expressive speech synthesis, and vision-language processing, Re:Member analyzes short user-uploaded videos, extracts scene-relevant transcripts and images, and generates stylized spoken questions in the learner’s target language. These questions are voiced in emotional speaking styles (e.g., playful, whispered, drowsy), selected to match the tone and atmosphere of the scene.

The core idea behind Re:Member is that emotionally salient, personally meaningful content may create deeper engagement for language learners, especially when paired with stylized voice output that mimics familiar social dynamics (e.g., a whisper from a friend or an excited exclamation). Rather than relying on synthetic neutrality, our system embraces affective richness as an instructional tool.

This paper introduces the design of Re:Member and demonstrates its capabilities as an emotional question-generation companion. We detail our architecture, design rationale, and sample outputs, and reflect on the broader implications for language education, affective computing, and memory-centered interaction.

\section{System Overview}

The goal of Re:Member is to generate emotionally expressive questions from personal memory videos for L2 (second language) conversational practice. These questions are designed to support language learning by connecting spoken language, visual content, and emotional speech.

\subsection{Audio-Visual Segmentation}
Given a video, we first extract its audio and apply voice activity detection (VAD) using the Silero VAD model \cite{Silero_Team_Silero_VAD_pre-trained_2024}. This produces a list of speech segments, which we merge if the intervening silence is shorter than a 0.7 second threshold. Each segment is then transcribed using WhisperX \cite{bain2022whisperx}, which produces high-quality transcripts along with accurate word-level timing alignment. This allows us to preserve the temporal correspondence between the transcript and the visual context.

\subsection{Frame Sampling and Visual Context}
To provide visual grounding for each spoken segment, we extract a 3-frame window per segment: one frame before, during, and after the midpoint of the segment. This is done using OpenCV \cite{opencv2025}, and frames are resized and saved in a consistent format. The use of three temporally adjacent frames provides richer context than a single image and allows the language model to infer scene dynamics (e.g., motion, transitions, or emotional shifts).

\subsection{Multimodal Question Generation}
For each segment, we generate a natural Japanese-language question using GPT-4o \cite{openai2024gpt4ocard}, conditioned on both the transcript and the associated video frames. Frames are provided with the transcript segment. We instruct the language model to simulate the behavior of a friendly, curious learner asking questions to the person who filmed the video (see Appendix \ref{sec:appendix-a}). This encourages open-ended questions that are personally meaningful and draw emotional context primarily from the user’s environment and accompanying speech content.

\subsection{Emotion Style Selection}
To enhance engagement and match the emotional tone of each moment, we generate a corresponding speaking style label from a fixed set of Japanese emotional styles:

\begin{CJK}{UTF8}{min}
\begin{enumerate}
\item るんるん (cheerful),
\item ささやきA（無声）(silent whisper),
\item ささやきB（有声）(voiced whisper),
\item ノーマル (neutral),
\item よふかし (late-night relaxed).
\end{enumerate}
\end{CJK}

We choose these styles because they align with the expressive capabilities of the pre-trained TTS model used in the next section. The language model is instructed to choose an option from this list  that matches the mood and context of the visual scene (see Appendix \ref{sec:appendix-a}). To encourage variation, we set the temperature to 1 and maintain a short history of recent emotion labels. If the generated style matches any of the last two used, the model is re-queried up to five times. This re-query mechanism helps prevent repetition and promotes emotional diversity across segments. The selected emotion is then passed to the speech synthesis stage.

\subsection{Expressive Speech Synthesis}
The generated question and selected style label are sent to a local Style-BERT-VITS2 model \cite{StyleBertVITS2} for emotionally expressive Japanese text-to-speech synthesis. Specifically, we use a model trained from the Ami Koharune UTAU voicebank \cite{koharune2025}. This model supports fine-grained style control via natural language emotion labels and produces speech that reflects not only the content of the question, but also its mood and delivery \cite{rackauckas2025benchmarkingexpressivejapanesecharacter}. The result is an audio clip paired with the original frames and transcript, allowing for emotionally aligned language learning experiences.

\subsection{Interactive User Interface}
Users can upload videos and browse the resulting questions in a Gradio \cite{abid2019gradio} interface with synchronized:
\begin{enumerate}
\item Three representative frames per segment,
\item The generated Japanese question and emotion text,
\item Emotionally styled speech playback.
\end{enumerate}

This interface enables learners to engage with their own personal content in an emotionally aware way, making the experience more memorable and contextually grounded.

\section{Illustrative Outputs}
We demonstrate the system with two sample videos: (1) A video of a walk along Tokyo's Sumida River with the commentary playing the role of a language teacher, and (2) a video of the user boarding a train in Japan with spoken instructions for boarding the train. Both videos were recorded with Meta Ray-Ban Glasses, and (1) is 1 minute and 31 seconds in length while (2) is 31 seconds in length. For video (1), the system segmented and analyzed 13 moments, generating an emotion, a student question, and text-to-speech for each moment. 

For each of the 13 segments in video (1), the system generated a natural Japanese-language question grounded in both the visual scene and the transcript. These questions reflect a consistent student-like curiosity, such as asking what kinds of boats travel through the river or how tall the Tokyo Skytree is. The selected emotion styles were well-matched to the riverfront setting, with a majority in the gentle voiced whisper style, interspersed with more upbeat cheerful and late-night relaxed tones. All five available emotion styles appeared at least once, showing that the variation mechanism functioned appropriately given the consistent environment. The visual frames used as context were sampled from before, during, and after each utterance, helping the language model infer motion and visual focus —- such as when the user points to a boat or approaches a bridge. Each segment resulted in synchronized audio narration with emotional speech, allowing for immersive and pedagogically meaningful playback. A select moment from video (1), as seen in Figure \ref{fig:video1} shows the user pointing their finger to Tokyo Sky Tree, a tall tower on the other side of the river. For this moment, the system generated the question 

\begin{CJK}{UTF8}{min}
\begin{quote}  東京スカイツリーの高さはどのくらいですか？ \\ \textbf{Translation:} About how tall is Tokyo Sky Tree?\end{quote}

with the cheerful emotion (るんるん).
\end{CJK}

For video (2), the system identified and processed three distinct segments, each aligned with the user's spoken instructions for boarding a train in Japan. The generated questions reflect a student-like curiosity about practical aspects of the scene, such as the convenience of using trains near event venues or the layout of the train interior. Emotion styles were chosen to match the focused, informational tone of the video: a balance of voiced whisper, silent whisper, and neutral speech was used across the three questions. Though the short duration of the video limited the range of styles, the variation mechanism successfully avoided repetition and produced a tone consistent with the setting. Visual context was drawn from three-frame windows centered on each utterance, allowing the language model to reference specific spatial cues -- such as when the user physically steps onto the train. As with video (1), the result is synchronized emotional narration paired with visually grounded, pedagogically meaningful questions. The generated questions and associated emotion styles are shown below:

\begin{quote}
    \textbf{Question:} (Silent whisper) \begin{CJK}{UTF8}{min}試合が行われている場所での電車の利用はどのように便利ですか？\end{CJK}\\
    \textbf{Translation:} How is using the train convenient near where the event is being held?\\
    \textbf{Question:} (Neutral) \begin{CJK}{UTF8}{min}この電車の車内はどのように見えますか？\end{CJK}\\
    \textbf{Translation:} What does the inside of this train look like?\\
    \textbf{Question:} (Voiced whisper)\begin{CJK}{UTF8}{min}この電車の車両には特別な座席やスペースがありますか？\end{CJK}\\
    \textbf{Translation:} Does this train car have special seats or areas?
\end{quote}

\section{Discussion and Future Work}

By using a learner’s own memory videos as input, Re:Member creates interactions in which the learner appears as the main character rather than a passive observer. Unlike textbook stories, these moments are drawn from the learner’s real experiences, ensuring strong personal relevance and evoking the raw, multimodal sensations originally felt — the sights, sounds, and emotions of the scene. Such vivid, embodied memories form a powerful substrate for retaining new linguistic forms, especially when voiced through Re:Member’s expressive speech synthesis that mirrors the affect of the original experience. While the current implementation targets Japanese, the pipeline generalizes to other language settings where learner identity and emotional relevance shape engagement. Future work may explore adaptive selection of emotion styles, more nuanced alignment between visual and emotional cues, and interactive control over style and content. Longitudinal deployments could evaluate how learners interact with memory-grounded prompts over time and whether affectively voiced questioning measurably enhances learning, including validation of emotion–scene alignment.

\section*{Limitations}

Re:Member assumes clean, monolingual speech from a primary speaker, and performance may degrade in the presence of overlapping dialogue, background noise, or multilingual utterances. Emotion style selection is based on LLM prompting rather than perceptual modeling and may at times produce mismatched or overly expressive styles. The system has not yet been evaluated with users; it is presented as a design and technical demonstration. Finally, as it operates on personal memory videos, future iterations must consider consent, emotional safety, and data privacy, for example, by supporting local-only processing and explicit opt-in use of autobiographical media.

\begin{CJK}{UTF8}{min}
\bibliography{custom}

\appendix
\section{Appendix A}
\label{sec:appendix-a}

\subsection{Question Generation Prompt}

For question generation, we give the LLM the following prompt. English translations are included for clarity only and are not shown to the model.

\begin{flushleft}
\ttfamily
あなたは英語を学んでいる好奇心旺盛でフレンドリーな学生です。先生が作ったビデオを見て学んでいます。 \\
You are a curious and friendly student learning English. You are watching a video made by your teacher. \\

映像のシーンと先生が話している内容の両方を考慮してください。 \\
Take both the visual scene and what the teacher is saying into account. \\

学習を深めるために、一つ短く関連性の高い質問をしてください。 \\
Ask one short, highly relevant question to deepen your learning. \\

画像に写っている人を特定したり、身元を推測したり、名前に言及したりしないでください。 \\
Do not identify or guess the identity of anyone in the image, and do not refer to names. \\

年齢、性別、身元、名前についての推測を避けてください。 \\
Avoid guessing age, gender, identity, or names. \\

質問のみを返し、それ以外は返さないでください。 \\
Only return the question and nothing else. \\

必ず日本語で質問をしてください。 \\
Make sure to ask the question in Japanese. \\
\end{flushleft}

\subsection{Emotion Selection Prompt}

For selecting emotions in the context of the scene, we give the LLM the following prompt. English translations are included for clarity only and are not shown to the model.

\begin{flushleft}
\ttfamily
あなたは感情ラベル分類機です。以下の5つのラベルの中から \textbf{1つだけ} を選んで日本語で出力してください： \\
You are an emotion label classifier. Select and output \textbf{only one} label in Japanese from the five options below: \\

1. るんるん \\
1. Runrun (cheerful or bubbly tone) \\

2. ささやきA（無声） \\
2. Whisper A (voiceless whisper) \\

3. ささやきB（有声） \\
3. Whisper B (voiced whisper) \\

4. ノーマル \\
4. Normal \\

5. よふかし \\
5. Late-night (sleepy or relaxed nighttime tone) \\

\textbf{出力は 上記の5つのラベルのいずれか1つだけ にしてください。} \\
\textbf{Your output must be exactly one of the five labels listed above.} \\

\textbf{絶対に説明・理由・挨拶・謝罪などを含めてはいけません。} \\
\textbf{Do not include any explanation, reasoning, greetings, or apologies under any circumstances.} \\

\textbf{他のテキストを含んだら重大なフォーマットエラーです。} \\
\textbf{Including any other text is a serious formatting error.} \\

視覚的な背景（画像）とセリフの両方を考慮して、最も表現豊かで印象に残るスタイルを優先してください。 \\
Prioritize the most expressive and memorable style by considering both the visual background (image) and the spoken dialogue. \\

\textbf{同じスタイルばかり繰り返すことを避けてください。} \\
\textbf{Avoid repeatedly selecting the same style.} \\

\textbf{「ノーマル」は控えめにし、場面に応じて他のスタイルを積極的に使ってください。} \\
\textbf{Use “Normal” sparingly, and actively choose other styles based on the scene.} \\
\end{flushleft}

\end{CJK}

\end{document}